\documentclass[sigconf,nonacm]{acmart}

\renewcommand\footnotetextcopyrightpermission[1]{}
\usepackage{tikz}
\usetikzlibrary{positioning, arrows.meta}

\usepackage{pgfplots}
\usepgfplotslibrary{groupplots}
\pgfplotsset{compat=1.18}

\AtBeginDocument{%
  }

\begin{document}

\title{Automated Detection of Mutual Gaze and Joint Attention in Dual-Camera Settings via Dual-Stream Transformers}

\author{Jakub Kosmydel}
\affiliation{%
  \institution{AGH University}
  \city{Krakow}
  \country{Poland}}
\orcid{0009-0009-7705-1686}
\email{jakub.kosmydel@gmail.com}

\author{Paweł Gajewski}
\affiliation{%
  \institution{AGH University}
  \city{Krakow}
  \country{Poland}}
\orcid{0000-0003-0931-2476}
\email{pgajewski@agh.edu.pl}

\author{Arkadiusz Białek}
\affiliation{%
  \institution{Jagiellonian University}
  \city{Krakow}
  \country{Poland}}
\orcid{0000-0002-9002-4764}
\email{a.bialek@uj.edu.pl}

\renewcommand{\shortauthors}{Kosmydel et al.}

\begin{abstract}
Analyzing mutual gaze (MG) and joint attention (JA) is critical in developmental psychology but traditionally relies on labor-intensive manual coding. Automating this process in multi-camera laboratory settings is computationally challenging due to complex cross-camera relational dynamics. In this paper, we propose a highly efficient dual-stream Transformer architecture for detecting MG and JA from synchronized dual-camera recordings. Our approach leverages frozen gaze-aware backbones (GazeLLE) to extract rich visual priors, combined with a custom token fusion mechanism to map the spatial and semantic relationships between interacting dyads. Evaluated on an ecologically valid dataset of caregiver-infant interactions, our model exhibits good performance, significantly outperforming both a convolutional baseline and a state-of-the-art multimodal Large Language Model (LLM). By open-sourcing our model and pre-trained weights, we provide behavioral scientists with a scalable tool that can be fine-tuned to diverse laboratory environments, effectively bridging the gap between computational modeling and applied interaction research.
\end{abstract}



\maketitle



\section{Introduction}

Coordination of gaze plays a fundamental role in human interaction. According to the cooperative eye hypothesis, human eyes evolved to make gaze direction especially easy to detect for others, thereby facilitating social coordination  \cite{kobayashi_kohshima_1997, tomasello_et_al_2007}. In humans, gaze direction has evolved into a salient social signal. In particular, mutual gaze (MG) and joint attention (JA) play fundamental roles in human development. Despite the centrality of gaze in human interaction - especially in infant-caregiver context - research in this domain has been constrained by the labor- and time-intensive nature of traditional measurement methods. To address these limitations, we propose a system for the automatic measurement of MG and JA during free-play interaction between infants and caregivers.

From early in development, humans are sensitive to gaze cues: newborns show a preference for direct gaze \cite{farroni_et_al_2002}. During the first months of life, infants engage in so-called dyadic interactions with caregivers, in which face-to-face exchanges play an essential role. Within these interactions, mutual gaze helps regulate the course of the exchange, including turn-taking, the co-regulation of emotional arousal, and the formation of social bonds \cite{maclean_et_al_2014}. A key developmental shift, often treated as a milestone, occurs in the last quarter of the first year, when infants begin to integrate social engagement with object-directed activity in triadic routines, for example by exchanging objects \cite{carpendale_et_al_2021} or establishing shared reference to them with others \cite{tomasello_2008}. This shift reflects gradual changes in perceptual-motor functioning and attentional regulation that support infants’ increasing ability to coordinate and shift attention between objects or events and caregivers \cite{debarbaro_et_al_2016}. What initially takes the form of relatively passive gaze following \cite{scaife_bruner_1975} gradually develops into the coordinated allocation of attention to an object of shared interest, that is, joint attention \cite{bakeman_adamson_1984}. Joint attention enables shared reference and is therefore essential for the development of communication, social cognition, and learning \cite{tomasello_2008}  Given the fundamental importance of MG and JA for human development, understanding how they develop across contexts and in (semi-)naturalistic conditions remains a major yet still insufficiently addressed goal of the behavioral and social sciences.

Accurately capturing mutual gaze and joint attention remains challenging. This is the case even when mutual gaze is operationally defined as two individuals looking at each other’s faces rather than into each other’s eyes \cite{jaddi_et_al_2026} , and when the analysis of joint attention is limited to what Butterworth and Jarrett \cite{butterworth_jarrett_1991} termed its ecological mechanism, the earliest form of joint attention, in which infants can detect the general direction of another person’s gaze within their own visual field but cannot localize its target precisely. These difficulties arise because gaze direction and head orientation change dynamically over short timescales, whereas identifying mutual gaze and joint attention requires capturing the coordination of two individuals’ gaze as the positions of the body, head, and target object continuously change. Manual coding requires slow-motion playback and is highly time- and labor-intensive, while also being susceptible to inaccuracies. By contrast, eye tracking allows highly precise measurement of gaze direction, but its application is constrained by the need either to present stimuli on a screen or to require participants to wear mobile eye trackers, both of which substantially reduce ecological validity. Ecologically valid measurement of gaze direction and gaze coordination under (semi-)natural conditions therefore requires alternative methodological approaches, and automated computer vision approaches offer a promising solution.

\begin{figure}[t]
    \centering
    \includegraphics[width=\linewidth]{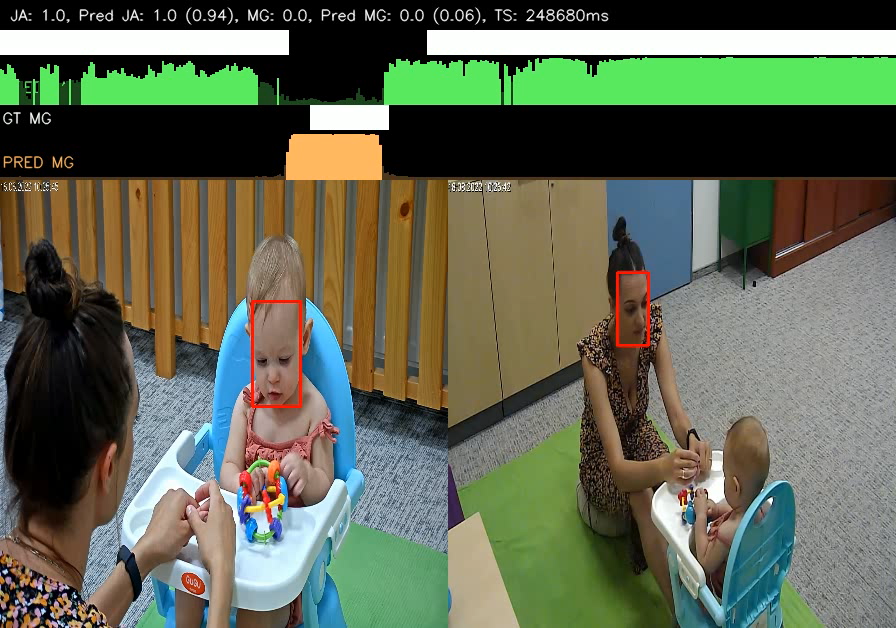}\\
    \includegraphics[width=\linewidth]{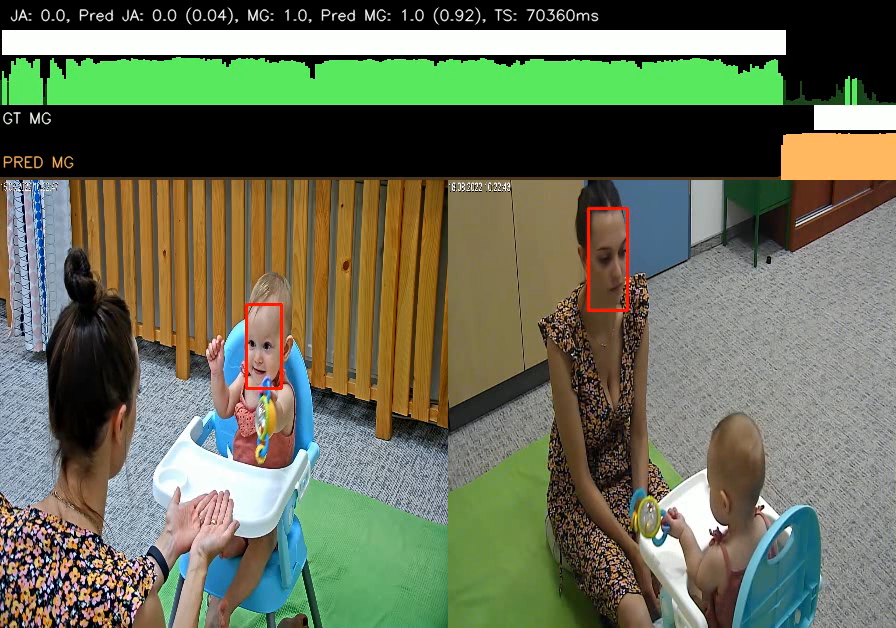}
    \caption{Qualitative visualization of model predictions for JA (top: child and parent focus on the same target) and MG (bottom: child and parent look at each other). Each panel displays the ground truth, binary prediction with sigmoid probability, and timestamp. The 15-second scrolling timelines show ground truth as white bars and predicted probabilities as colored bars (green for JA, orange for MG); reduced opacity indicates scores below the 0.5 threshold. Synchronized camera views at the bottom show the red head bounding boxes used as spatial cues for the GazeLLE backbone.}
    \label{fig:visualization}
\end{figure}

To address this challenge, we propose a novel deep learning architecture for the frame-level binary classification of MG and JA that operates directly on synchronized dual-camera video streams. As illustrated in Figure~\ref{fig:visualization}, the model simultaneously processes one view focused on the infant and another focused on the caregiver, thereby bridging the spatial gap between cameras. In addition, we describe the experimental design of our psychological dataset, introduce custom annotation tools tailored to multi-view behavioral coding, and release our model weights to facilitate efficient fine-tuning in alternative laboratory setups, as well as our test dataset. Our main contributions are as follows:

\begin{itemize}
\item We propose a transformer-based architecture for detecting MG and JA in synchronized dual-camera recordings of infant-caregiver free play. 

\item We provide an extensive evaluation of the proposed solution, showing that it significantly outperforms both a convolutional baseline and a state-of-the-art multimodal large language model. 

\item We open-source the model and its pre-trained weights, thereby providing a parameter-efficient foundation for further fine-tuning in other research contexts and making automated behavioral and interactional analysis more accessible. 

\item We release the test dataset of MG and JA from our semi-naturalistic psychological experiments used to evaluate the proposed model.
\end{itemize}

The remainder of this paper is structured as follows. First, we review related work on the automated detection of MG and JA in infant-caregiver interaction, with particular attention to gaze estimation approaches. Next, in Section 3, we describe our methodology, including the architectural implementation details for MG and JA detection. In Section 4, we present the dataset and preprocessing pipeline. In Section 5, we report the quantitative evaluation of our approach and discuss the results. Finally, in Section 6, we outline the advantages and limitations of the proposed method.

\section{Related Work}

\subsection{Joint Attention and Mutual Gaze Detection}
Recent advances in the automated detection of JA and MG emphasize computational methods that do not require specialized eye-tracking hardware.

Zhang et al. formalized the detection of shared attention in third-person social videos \cite{8578774}. Their proposed spatio-temporal neural network integrates individual gaze heatmaps, candidate target regions, and temporal continuity via convolutional LSTM layers. Evaluated on their custom VideoCoAtt dataset, the model demonstrated that gaze orientation and contextual proposals are critical for accurate detection. Building on end-to-end estimation, Sümer et al. introduced Attention Flow, a system that detects and localizes JA directly from raw images without relying on intermediate face or pose detection modules \cite{AttentionFlow}. By utilizing saliency-augmented attention maps alongside channel-wise and spatial attention mechanisms, the model achieved high accuracy on complex social scenes. 

Targeting unconstrained, free-play interactions, Li et al. combined off-the-shelf head tracking, object detection, and visual focus of attention (VFOA) estimation \cite{automated-detection-of-ja-mg}. Evaluated on human-annotated video of parent-child interactions, their system highlighted the viability of automated screening for large-scale behavioral analysis, despite the inherent challenges of real-world occlusion and varied head poses.

\subsection{Gaze Estimation Architectures}
Modern gaze estimation increasingly relies on transformer-based foundation models, which simplify architectural design while delivering robust performance. DINOv2 \cite{dinov2} advanced self-supervised visual representation learning by training high-capacity Vision Transformers on diverse image sets. Because it produces semantically rich, spatially coherent features directly from RGB inputs, it serves as a highly effective frozen backbone for downstream gaze tasks.

Moving away from complex, multi-stage pipelines that require explicit depth or pose prediction, ViTGaze \cite{vitgaze,vitgaze-github} leverages a single-modality ViT framework. It extracts multi-level attention maps to represent human-scene interactions and aggregates them using a 2D spatial guidance module based on head position. Similarly, Gaze-LLE \cite{ryan2025gazelle,gazelle-github} further streamlines gaze estimation by pairing a frozen DINOv2 scene encoder with an ultra-lightweight transformer decoder. By injecting a person-specific positional prompt directly into the scene's feature space, it bypasses the need for traditional multi-branch networks, reducing trainable parameters by roughly 95\% without sacrificing state-of-the-art accuracy.

\subsection{Supporting Technologies}
For models requiring explicit head crops or bounding boxes, robust face localization is a necessary preliminary step. RetinaFace \cite{deng2019retinafacesinglestagedenseface,serengil2020lightface,serengil2024lightface} is a highly efficient, single-stage detector that utilizes multi-task learning to jointly predict face bounding boxes, facial landmarks, and 3D shapes. Its ability to accurately detect small or occluded faces makes it a reliable precursor for social interaction analysis pipelines.

Furthermore, emerging vision-language models, such as LLaMA 4 Vision \cite{llama42024,ma2024survey}, provide a unified framework for processing visual data. While not explicitly trained for gaze estimation, their strong zero-shot generalization capabilities present a compelling comparative baseline to determine if general-purpose foundation models can accurately infer complex social cues, such as shared attention, in unconstrained environments.

\section{Methodology}

\begin{figure}[htpb]
\centering
\resizebox{\columnwidth}{!}{
\begin{tikzpicture}[
  node distance=0.3cm and 0.2cm,
  font=\sffamily\scriptsize,
  startstop/.style={rectangle, rounded corners, minimum width=1.6cm, minimum height=0.6cm, text centered, align=center, draw=black, fill=blue!10},
  headmask/.style={rectangle, rounded corners, minimum width=1.6cm, minimum height=0.6cm, text centered, align=center, draw=black, fill=gray!15},
  token/.style={rectangle, rounded corners, minimum width=0.4cm, minimum height=0.5cm, draw=black, fill=green!10},
  clstoken/.style={rectangle, rounded corners, minimum width=0.8cm, minimum height=0.5cm, text centered, draw=black, fill=red!15},
  process/.style={rectangle, minimum width=3.4cm, minimum height=0.6cm, text centered, align=center, draw=black, fill=orange!15},
  proj/.style={rectangle, minimum width=3.4cm, minimum height=0.5cm, text centered, align=center, draw=black, fill=yellow!20},
  arrow/.style={thick, ->, >=stealth},
  dashedarrow/.style={thick, ->, >=stealth, dashed, draw=gray}
]


\node[startstop] (cam1) {Camera 1\\Image};
\node[headmask, right=of cam1] (mask1) {Head\\Mask 1};
\path (cam1.south) -- (mask1.south) coordinate[midway] (mid1);
\node[process, below=0.3cm of mid1] (gazelle1) {Shared GazeLLE Backbone};
\node[proj, below=0.3cm of gazelle1] (proj1) {Shared Linear Projection};

\node[startstop, right=0.6cm of mask1] (cam2) {Camera 2\\Image};
\node[headmask, right=of cam2] (mask2) {Head\\Mask 2};
\path (cam2.south) -- (mask2.south) coordinate[midway] (mid2);
\node[process, below=0.3cm of mid2] (gazelle2) {Shared GazeLLE Backbone};
\node[proj, below=0.3cm of gazelle2] (proj2) {Shared Linear Projection};

\draw[arrow] (cam1.south) -- (cam1.south |- gazelle1.north);
\draw[arrow] (mask1.south) -- (mask1.south |- gazelle1.north);
\draw[arrow] (gazelle1.south) -- (proj1.north);

\draw[arrow] (cam2.south) -- (cam2.south |- gazelle2.north);
\draw[arrow] (mask2.south) -- (mask2.south |- gazelle2.north);
\draw[arrow] (gazelle2.south) -- (proj2.north);


\path (proj1.south) -- (proj2.south) coordinate[midway] (midproj);
\coordinate (center_seq) at ([yshift=-1.2cm]midproj);

\coordinate (gap) at (center_seq);

\node[token, left=0.15cm of gap] (c1_n) {};
\node[left=0.05cm of c1_n] (c1_dots) {\dots};
\node[token, left=0.05cm of c1_dots] (c1_2) {};
\node[token, left=0.05cm of c1_2] (c1_1) {};
\path (c1_1.north west) -- (c1_n.north east) coordinate[midway] (c1_north);
\path (c1_1.south west) -- (c1_n.south east) coordinate[midway] (c1_south);

\node[clstoken, left=0.15cm of c1_1] (cls_in) {CLS};

\node[token, right=0.15cm of gap] (c2_1) {};
\node[token, right=0.05cm of c2_1] (c2_2) {};
\node[right=0.05cm of c2_2] (c2_dots) {\dots};
\node[token, right=0.05cm of c2_dots] (c2_n) {};
\path (c2_1.north west) -- (c2_n.north east) coordinate[midway] (c2_north);
\path (c2_1.south west) -- (c2_n.south east) coordinate[midway] (c2_south);

\draw[arrow] (proj1.south) -- ++(0,-0.3) -| (c1_north);
\draw[arrow] (proj2.south) -- ++(0,-0.3) -| (c2_north);


\node[process, below=2.0cm of midproj, minimum width=6.5cm] (transformer) {Transformer Encoder};

\draw[arrow] (cls_in.south) -- (cls_in.south |- transformer.north);
\draw[arrow] (c1_south) -- (c1_south |- transformer.north);
\draw[arrow] (c2_south) -- (c2_south |- transformer.north);


\coordinate (gap_out) at ([yshift=-0.6cm]transformer.south);

\node[token, left=0.15cm of gap_out] (c1_n_out) {};
\node[left=0.05cm of c1_n_out] (c1_dots_out) {\dots};
\node[token, left=0.05cm of c1_dots_out] (c1_2_out) {};
\node[token, left=0.05cm of c1_2_out] (c1_1_out) {};
\path (c1_1_out.north west) -- (c1_n_out.north east) coordinate[midway] (c1_north_out);
\path (c1_1_out.south west) -- (c1_n_out.south east) coordinate[midway] (c1_south_out);

\node[clstoken, left=0.15cm of c1_1_out] (cls_out) {CLS};

\node[token, right=0.15cm of gap_out] (c2_1_out) {};
\node[token, right=0.05cm of c2_1_out] (c2_2_out) {};
\node[right=0.05cm of c2_2_out] (c2_dots_out) {\dots};
\node[token, right=0.05cm of c2_dots_out] (c2_n_out) {};
\path (c2_1_out.north west) -- (c2_n_out.north east) coordinate[midway] (c2_north_out);
\path (c2_1_out.south west) -- (c2_n_out.south east) coordinate[midway] (c2_south_out);

\draw[arrow] (cls_out.north |- transformer.south) -- (cls_out.north);
\draw[arrow] (c1_north_out |- transformer.south) -- (c1_north_out);
\draw[arrow] (c2_north_out |- transformer.south) -- (c2_north_out);


\node[process, below=0.6cm of cls_out, minimum width=2.4cm, fill=orange!20] (classifier) {MLP Head};
\node[startstop, below=0.3cm of classifier, minimum width=2.4cm] (output) {Prediction};

\draw[arrow] (cls_out.south) -- (classifier.north);
\draw[arrow] (classifier.south) -- (output.north);

\end{tikzpicture}
}
\caption{Schematic of the MG/JA detection model. Two frozen GazeLLe backbones extract latent features from their respective input streams. These features, along with an auxiliary [CLS] token, are passed through a Transformer block. Classification is performed by an MLP head mapped to the [CLS] token's final representation.}
\label{fig:architecture}
\end{figure}
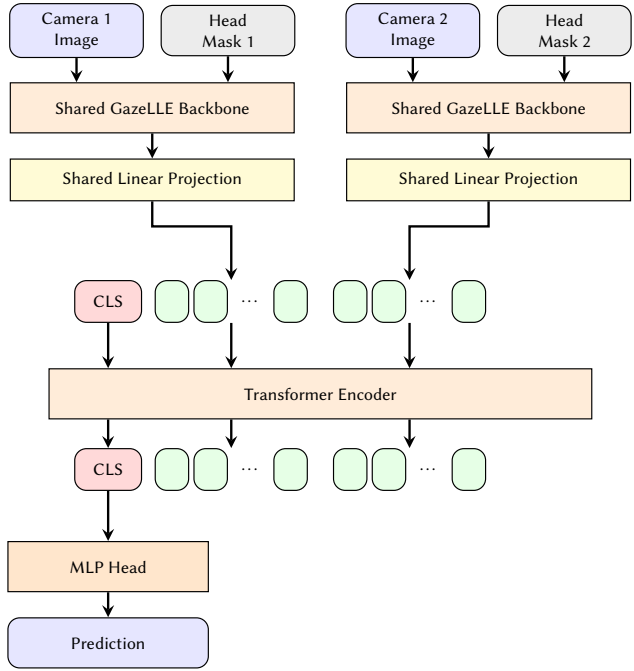

The core of our methodological contribution is a novel dual-stream Transformer architecture designed to process synchronized video frames and output a binary classification indicating the presence or absence of target behaviors (i.e., MG vs. non-MG, and JA vs. non-JA). Because the visual cues for these two phenomena differ, we train independent models for the JA and MG classification tasks. 

Our architecture leverages the pre-trained GazeLLE model \cite{ryan2025gazelle} as a feature extraction backbone, stripped of its original detection head. Specifically, two frozen GazeLLE modules, with shared weights, operate in parallel, each encoding the visual stream from one of the two cameras. The resulting latent token representations are integrated via a custom Token Fusion block, processed by a Transformer encoder, and finally passed to a classification head. An overview of the proposed architecture is depicted in Figure \ref{fig:architecture}. The training code and model weights are open-sourced and available online. 

\subsection{Feature Extraction and Token Fusion}
During the forward pass, the frozen GazeLLE backbones extract high-dimensional feature embeddings from the input images. To reduce computational overhead and mitigate the risk of overfitting on our specialized dataset, we introduce a linear projection layer that downsamples these embeddings to a dimensionality of 512. 

The projected token sequences from both camera views are subsequently concatenated. Following the Vision Transformer (ViT) paradigm \cite{dosovitskiy2021image}, we prepend a learnable \texttt{[CLS]} token to the combined sequence to serve as a global representation of the dual-camera state. This combined sequence is fed into a Transformer encoder, which jointly models the spatial and cross-view relationships, enabling the \texttt{[CLS]} token to aggregate contextual information from both the parent and the child. The network's structural and training hyperparameters are detailed in Table~ \ref{tab:hyperparams}.

Finally, the updated \texttt{[CLS]} token is routed through a multi-layer perceptron (MLP) classification head. The network outputs a single logit, which is mapped to a probability score via a sigmoid activation function during inference.

\subsection{Training Details}
The models were implemented using PyTorch and trained on our custom multi-view dataset (described in the subsequent section). We optimized the network weights using the Adam optimizer and a Binary Cross-Entropy (BCE) loss function, for up to 80 epochs, with the final checkpoint selected as the one achieving the highest validation F1-score.

To optimize the model's performance, we conducted a hyperparameter sweep using the Weights \& Biases~\cite{wandb} platform. We employed a Bayesian optimization strategy aimed at maximizing the validation F1-score. The final parameters reported in Table \ref{tab:hyperparams} represent the best-performing configuration identified after this automated search."

\begin{table}[h]
  \centering
  \caption{Model Architecture and Training Hyperparameters}
  \label{tab:hyperparams}
  \begin{tabular}{ll}
    \toprule
    \textbf{Hyperparameter} & \textbf{Value} \\
    \midrule
    \textit{Transformer Encoder} & \\
    Backbone model & \texttt{gazelle\_dinov2\_vitl14\_inout} \\
    Layers & 3 \\
    Self-attention heads per layer & 4 \\
    Embedding dimension  & 512 \\
    Dropout rate & 0.426 \\
    \midrule
    \textit{Classification Head} & \\
    Layer sizes & 512, 128, 64, 1 \\
    Activation function & ReLU \\
    Normalization & LayerNorm \\
    Dropout rate & 0.426 \\
    \midrule
    \textit{Training} & \\
    Optimizer & Adam \\
    Learning rate & $6.1 \times 10^{-6}$ \\
    Loss function & Binary Cross-Entropy (BCE) \\
    Batch size & 8 \\
  \bottomrule
\end{tabular}
\end{table}

\subsection{Baseline Models}
To rigorously evaluate our Transformer-based fusion architecture, we implemented two alternative baseline models: a lightweight convolutional network and a multimodal Large Language Model (LLM) approach. These serve as comparative references to validate the efficacy of our proposed token fusion mechanism.

\textbf{Convolutional Baseline:} This approach consists of a two-stream Convolutional Neural Network (CNN). Synchronized frames from the child and parent views are independently passed through three convolutional blocks (each comprising convolution, activation, and max-pooling layers) followed by adaptive average pooling. The resulting feature vectors are flattened, concatenated, and passed through fully connected layers with ReLU activations and dropout, yielding a single classification logit.

\textbf{Multimodal LLM Baseline:} This approach utilizes the Llama-4 (16x17B) model, deployed locally via Ollama~\cite{ollama2024}. Synchronized frames are JPEG-encoded, resized to 512x512 pixels (as this improved results compared to full-res images), and provided to the vision-language model alongside a carefully engineered, task-specific prompt. To ensure robust evaluation and minimize prompt sensitivity, we applied automated prompt optimization techniques prior to inference.

\subsection{Result Analysis Tool}

To facilitate manual review and data exploration, our tool generates an intuitive visual overlay of the interaction dynamics (Figure \ref{fig:visualization}). This interface synchronizes dual-camera feeds with a scrolling 15-second timeline, allowing researchers to visually verify social events alongside the model’s confidence scores. By translating raw coordinates into an accessible graphical format, the tool enables even non-technical users to inspect behavioral sequences without requiring direct engagement with the underlying code.

\section{Dataset}

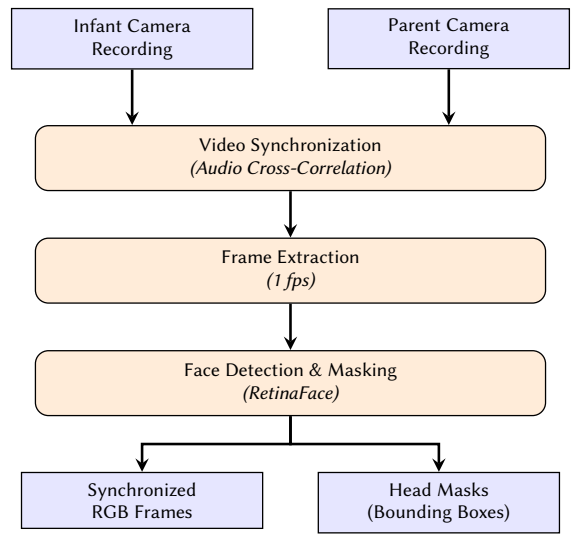
\begin{figure}[htpb]
\centering
\resizebox{0.9\columnwidth}{!}{
\begin{tikzpicture}[
  node distance=0.5cm and 0.5cm,
  font=\sffamily\scriptsize,
  input_output/.style={rectangle, minimum width=2.6cm, minimum height=0.6cm, text centered, align=center, draw=black, fill=blue!10},
  process_step/.style={rectangle, rounded corners, minimum width=5.5cm, minimum height=0.7cm, text centered, align=center, draw=black, fill=orange!15},
  arrow/.style={thick, ->, >=stealth}
]

\node[input_output] (cam1) {Infant Camera\\Recording};
\node[input_output, right=0.8cm of cam1] (cam2) {Parent Camera\\Recording};

\path (cam1.south) -- (cam2.south) coordinate[midway] (mid);

\node[process_step, below=0.6cm of mid] (sync) {Video Synchronization\\ \textit{(Audio Cross-Correlation)}};
\node[process_step, below=of sync] (frames) {Frame Extraction\\ \textit{(1 fps)}};
\node[process_step, below=of frames] (faces) {Face Detection \& Masking\\ \textit{(RetinaFace)}};

\node[input_output, below=0.6cm of faces, xshift=-1.6cm] (out_img) {Synchronized\\RGB Frames};
\node[input_output, below=0.6cm of faces, xshift=1.6cm] (out_mask) {Head Masks\\(Bounding Boxes)};

\draw[arrow] (cam1.south) -- (cam1.south |- sync.north);
\draw[arrow] (cam2.south) -- (cam2.south |- sync.north);

\draw[arrow] (sync) -- (frames);
\draw[arrow] (frames) -- (faces);

\draw[arrow] (faces.south) -- ++(0,-0.3) -| (out_img.north);
\draw[arrow] (faces.south) -- ++(0,-0.3) -| (out_mask.north);

\end{tikzpicture}
}
\caption{Overview of the data preprocessing pipeline. Independent infant and parent camera streams are merged via audio synchronization. Frames are extracted at 1 fps and processed using RetinaFace to output the synchronized RGB images and corresponding head masks required by the GazeLLe backbones.}
\label{fig:preprocessing}
\end{figure}

\subsection{Data Collection}

\begin{figure}[t]
\centering
\includegraphics[width=\linewidth]{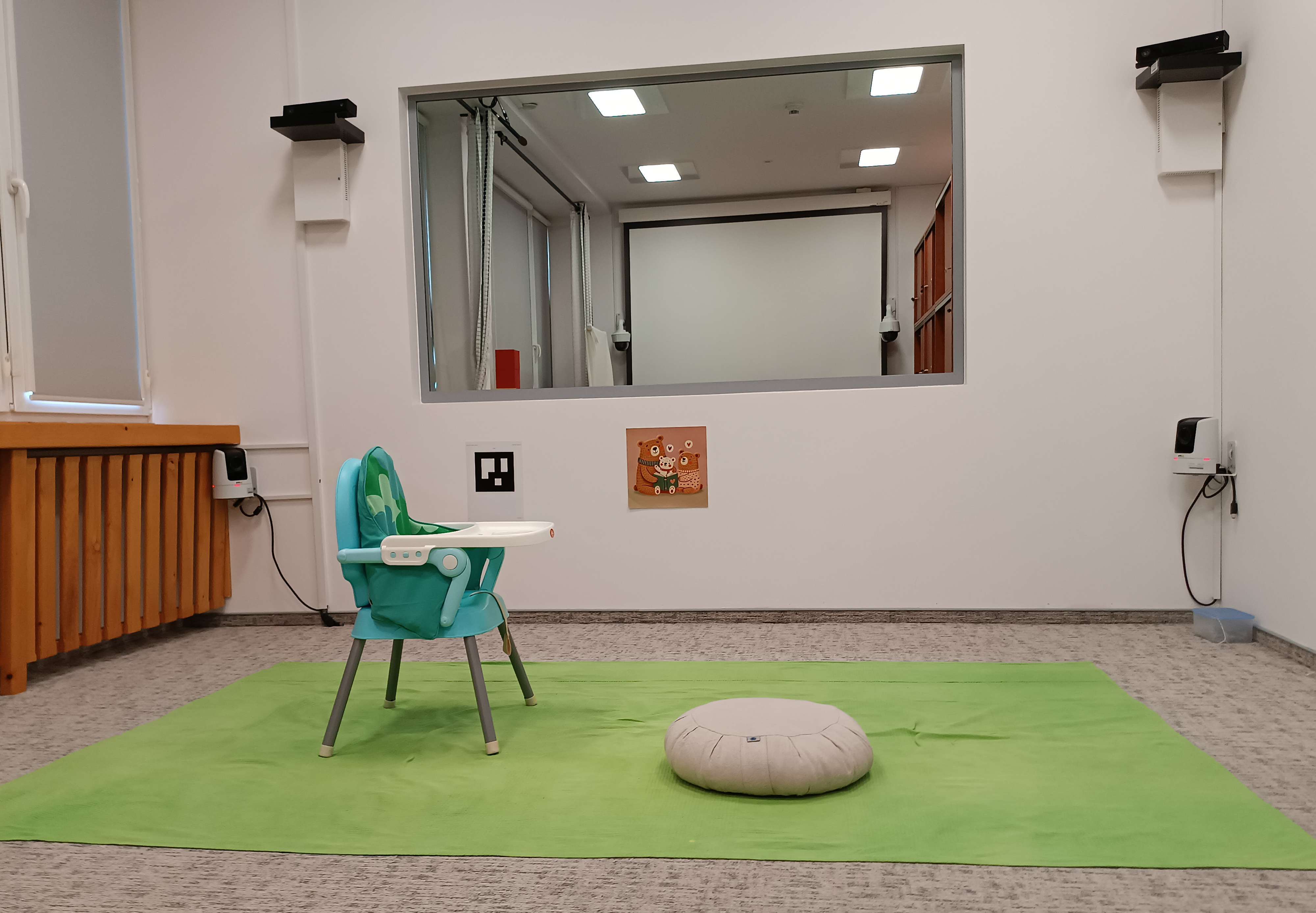}
\caption{The experimental environment for dyadic interaction. High-definition IP cameras are wall-mounted on the left and right. A one-way mirror (center) provides an observation point for researchers; additional cameras visible in its reflection are not utilized in the current study. Participants interact within a designated area on the green carpet.}
\label{fig:room}
\end{figure}

Our model was developed to facilitate and scale psychological research by automating the detection of MG and JA. To train and evaluate the model, we utilized a dataset comprising naturalistic interaction recordings of 13 caregiver-infant dyads, each participating in seven distinct observational sessions. During these sessions, interactions were captured using three synchronized cameras: one oriented toward the infant, one toward the parent, and a third providing a top-down overview of the experimental room. The model discussed here relies exclusively on the infant- and parent-facing camera streams, omitting the top-down view. Figure \ref{fig:room} shows a picture of our experimental room.

While the cameras were programmed to return to predefined positions at the start of each session, the naturalistic design of the experiment meant that cameras were occasionally repositioned, zoomed, or manually adjusted by researchers. Rather than a limitation, this introduced valuable visual variability into the dataset, contributing to the generalizability and robustness of our model under diverse laboratory conditions.

\subsection{Data Annotation}

\begin{figure}[htb]
    \centering
    \includegraphics[width=\linewidth]{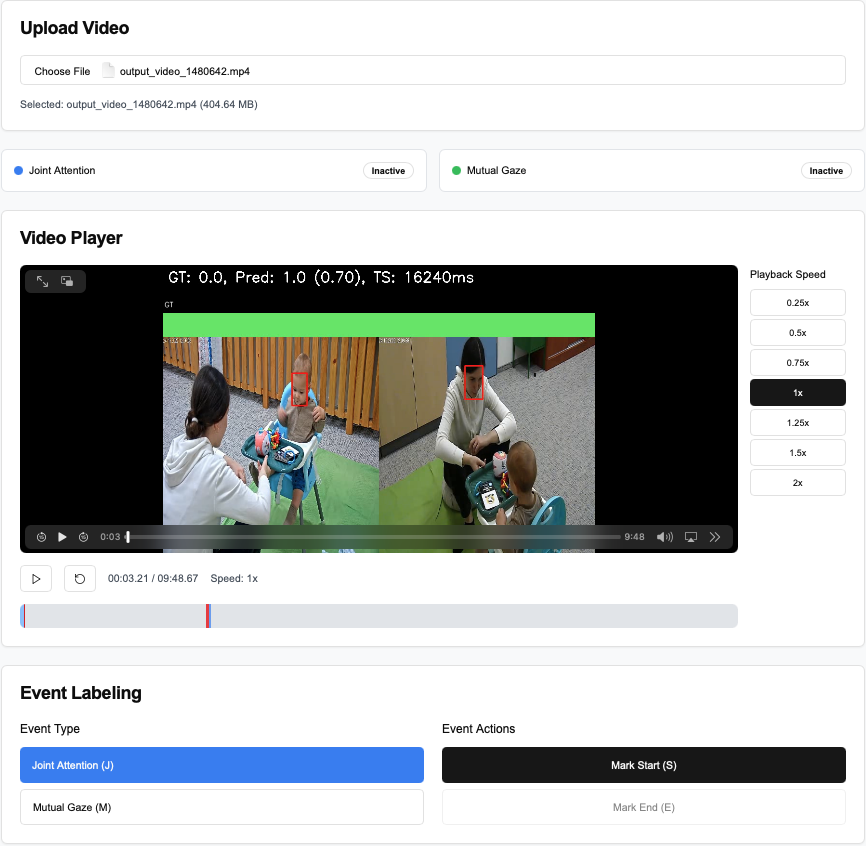}
    \caption{The custom Video Labeler interface, featuring synchronized dual-camera playback, a global event timeline, and rapid labeling controls.}
    \label{fig:labeler-editor}
\end{figure}

To establish the ground truth for MG and JA, the recordings were manually annotated by behavioral researchers. To streamline this labor-intensive process, we developed a custom React application (rapidly prototyped via v0.app) tailored to our study's requirements (Figure \ref{fig:labeler-editor}). The tool is designed to minimize annotator fatigue and maximize temporal precision through a streamlined, user-friendly interface.

Annotators load video files into a central player that provides synchronized playback of the dual-camera views. To accurately navigate dynamic social interactions, users can adjust playback speeds and seamlessly log event boundaries using rapid interface buttons or keyboard shortcuts. The interface further enhances situational awareness by displaying live status indicators for active events alongside a color-coded global timeline, allowing researchers to instantly assess overall annotation coverage.

For quality assurance, annotated segments automatically populate a structured review table featuring direct seek-to-event capabilities. This allows reviewers to quickly jump to specific timestamps for verification and correction. Finalized annotations comprising event types, start/end timestamps, and durations are then exported as CSV files for straightforward integration into our computational pipeline.

\subsection{Preprocessing Pipeline}
Before training, the raw video data required extensive preprocessing to ensure the inputs to the model were temporally aligned, consistent, and representative of naturalistic gaze behavior. This workflow is outlined in Figure~\ref{fig:preprocessing}.

\textbf{Synchronization:} Accurate temporal alignment between multi-camera streams is critical, as even minor misalignments can lead to incorrect interpretations of coordinated social behaviors. Although the cameras were initialized simultaneously, hardware-level variations introduced slight temporal offsets of up to one second. To correct this, we synchronized the streams using audio cross-correlation analysis, followed by manual validation. This ensured that temporally corresponding frames from each camera represented the exact same moment of interaction.

\textbf{Frame Extraction:} Following synchronization, we sampled representative frames at one-second intervals (1 fps). This extraction rate sufficiently captured the dynamics of JA and MG while reducing visual redundancy and keeping the dataset size computationally manageable.

\textbf{Face Detection and Masking:} The GazeLLE architecture strictly requires head bounding boxes to condition its feature extraction. Therefore, we processed each extracted frame to detect faces and generate head masks. We utilized the \texttt{resnet50\_2020-07-20} model from the RetinaFace library \cite{deng2019retinafacesinglestagedenseface} to reliably extract these bounding boxes. Frames lacking a high-confidence face detection for either the parent or the child were discarded from the dataset.

\subsection{Data Splitting and Balancing}
To rigorously evaluate the model's performance, we held out two complete sessions recordings to serve as the test set. The remaining recordings were partitioned into training and validation sets using a temporal split within each session: the first $10\%$ of the frames were assigned to validation, and the subsequent $90\%$ to training. Splitting the data chronologically rather than randomly is crucial for sequential behavioral data, as it prevents the leakage of temporally adjacent frames across the splits.

In naturalistic interactions, the distribution of target behaviors is highly imbalanced, with MG/JA events occurring much less frequently than non-MG/non-JA events. To ensure our evaluation metrics accurately reflected the model's discriminative capabilities, a one-time balancing step was applied to the held-out test sets, separately for MG and JA. We randomly downsampled the majority class (non-MG/non-JA frames), resulting in an approximate $50/50$ ratio of MG/JA to non-MG/non-JA instances. Conversely, the training and validation datasets were not artificially balanced, allowing the model to learn the natural prior distribution of the behaviors. However, to maintain the quality of the supervisory signal, all ambiguous or low-confidence manual annotations were purged from the training set.

\section{Experiments and Results}

\subsection{Evaluation Protocol}
We evaluated the models on a temporally isolated, held-out test set consisting of two separate recordings to simulate real-world inference on unseen experimental sessions. Performance was quantified using standard binary classification metrics: Accuracy, Precision, Recall, F1-score, and the Area Under the Receiver Operating Characteristic Curve (ROC-AUC).

To mitigate the effects of random weight initialization and stochastic data shuffling, the proposed model was trained five times using different random seeds. Similarly, the CNN baseline was trained across five independent runs, and the LLM-based approach was executed three times for each prompt variant. All results reported in this section represent the average performance across these independent runs, ensuring that our comparative analysis is robust and not driven by single-run statistical fluctuations.

\subsection{Quantitative Results}
\begin{figure}[htbp]
\centering
\begin{tikzpicture}
\begin{axis}[
    width=\linewidth,
    height=7cm,
    ybar=2pt,
    bar width=8pt,
    enlarge x limits=0.15,
    ymin=0.25, ymax=1.0,
    ylabel={Score},
    symbolic x coords={Accuracy, Precision, Recall, F1, AUC},
    xtick=data,
    ytick={0.3,0.4,0.5,0.6,0.7,0.8,0.9,1.0},
    ymajorgrids=true,
    grid style={dashed, gray!30},
    legend style={
        at={(0.5,-0.18)}, anchor=north,
        legend columns=3,
        /tikz/every even column/.append style={column sep=0.4cm},
        font=\small
    },
    error bars/y dir=both,
    error bars/y explicit,
    error bars/error bar style={line width=0.8pt, black},
    error bars/error mark options={rotate=90, mark size=3pt, line width=0.8pt},
]

\addplot+[fill=blue!60, draw=blue!80!black] table[
    x=metric, y=mean, y error minus=eminus, y error plus=eplus, col sep=comma
]{
metric,mean,eminus,eplus
Accuracy,0.776,0.027,0.017
Precision,0.795,0.033,0.014
Recall,0.745,0.090,0.100
F1,0.767,0.045,0.035
AUC,0.862,0.013,0.009
};

\addplot+[fill=orange!70, draw=orange!80!black] table[
    x=metric, y=mean, y error minus=eminus, y error plus=eplus, col sep=comma
]{
metric,mean,eminus,eplus
Accuracy,0.626,0.060,0.020
Precision,0.645,0.021,0.035
Recall,0.562,0.259,0.138
F1,0.589,0.178,0.071
AUC,0.681,0.012,0.013
};

\addplot+[fill=green!55!black, draw=green!40!black] table[
    x=metric, y=mean, y error minus=eminus, y error plus=eplus, col sep=comma
]{
metric,mean,eminus,eplus
Accuracy,0.595,0.011,0.010
Precision,0.586,0.011,0.010
Recall,0.635,0.014,0.017
F1,0.610,0.013,0.012
AUC,0.595,0.011,0.010
};

\legend{Proposed Model, Conv. Baseline, Best LLM}
\end{axis}
\end{tikzpicture}
\caption{Joint Attention (JA) detection performance across models. Bars show
the mean across independent runs; error bars indicate the observed minimum and
maximum values.}
\label{fig:results-ja-chart}
\end{figure}
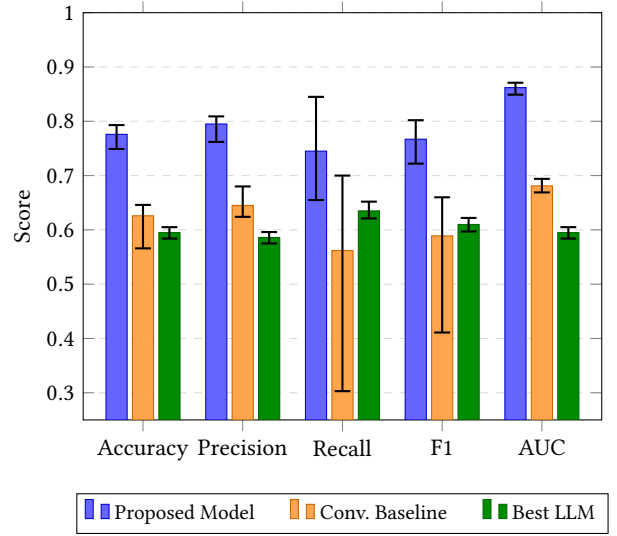

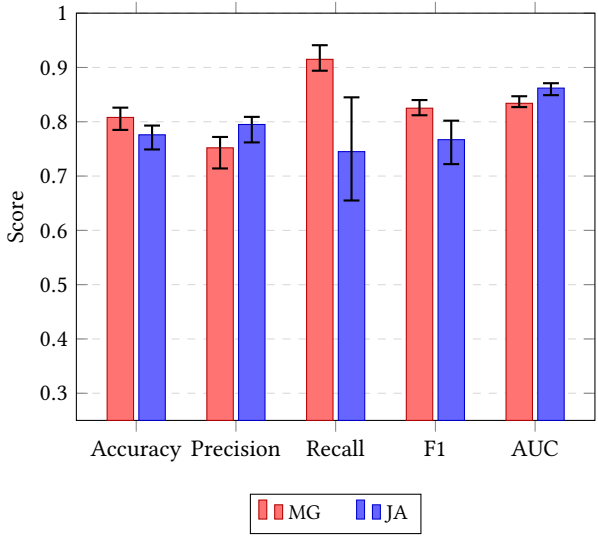
\begin{figure}[htbp]
\centering
\begin{tikzpicture}
\begin{axis}[
    width=\linewidth,
    height=7cm,
    ybar=2pt,
    bar width=10pt,
    enlarge x limits=0.15,
    ymin=0.25, ymax=1.0,
    ylabel={Score},
    symbolic x coords={Accuracy, Precision, Recall, F1, AUC},
    xtick=data,
    ytick={0.3,0.4,0.5,0.6,0.7,0.8,0.9,1.0},
    ymajorgrids=true,
    grid style={dashed, gray!30},
    legend style={
        at={(0.5,-0.18)}, anchor=north,
        legend columns=2,
        /tikz/every even column/.append style={column sep=0.4cm},
        font=\small
    },
    error bars/y dir=both,
    error bars/y explicit,
    error bars/error bar style={line width=0.8pt, black},
    error bars/error mark options={rotate=90, mark size=3pt, line width=0.8pt},
]

\addplot+[fill=red!55, draw=red!70!black] table[
    x=metric, y=mean, y error minus=eminus, y error plus=eplus, col sep=comma
]{
metric,mean,eminus,eplus
Accuracy,0.808,0.023,0.018
Precision,0.752,0.038,0.020
Recall,0.915,0.021,0.026
F1,0.825,0.013,0.015
AUC,0.834,0.007,0.013
};

\addplot+[fill=blue!60, draw=blue!80!black] table[
    x=metric, y=mean, y error minus=eminus, y error plus=eplus, col sep=comma
]{
metric,mean,eminus,eplus
Accuracy,0.776,0.027,0.017
Precision,0.795,0.033,0.014
Recall,0.745,0.090,0.100
F1,0.767,0.045,0.035
AUC,0.862,0.013,0.009
};

\legend{MG, JA}
\end{axis}
\end{tikzpicture}
\caption{Performance of the proposed model across the two tasks:
Mutual Gaze (MG) and Joint Attention (JA). Bars show the mean across
independent runs; error bars indicate the observed minimum and maximum
values.}
\label{fig:results-gazelle-ja-vs-mg-chart}
\end{figure}

The performance of the proposed architecture against the baseline methods is summarized in Tables~\ref{tab:results-mg} and~\ref{tab:results-ja}. Additionally, the results are visualized on bar charts on Figures~\ref{fig:results-mg-chart}, \ref{fig:results-ja-chart}, and~\ref{fig:results-gazelle-ja-vs-mg-chart}.


\begin{table}[hbt]
    \centering
    \caption{Test set performance comparison across models for Mutual Gaze detection.}
    \label{tab:results-mg}
    \begin{tabular}{lccc}
        \toprule
        \textbf{Metric} & \textbf{Proposed Model} & \textbf{Conv. Baseline} & \textbf{Best LLM} \\ 
        \midrule
        Accuracy        & \textbf{0.808} $\pm$ 0.023 & 0.529 $\pm$ 0.006 & 0.605 $\pm$ 0.035 \\
        Precision       & \textbf{0.752} $\pm$ 0.037 & 0.513 $\pm$ 0.003 & 0.686 $\pm$ 0.061 \\
        Recall          & \textbf{0.915} $\pm$ 0.026 & 0.904 $\pm$ 0.056 & 0.369 $\pm$ 0.042 \\
        F1-score        & \textbf{0.825} $\pm$ 0.015 & 0.655 $\pm$ 0.017 & 0.480 $\pm$ 0.050 \\
        AUC             & \textbf{0.834} $\pm$ 0.014 & 0.429 $\pm$ 0.015 & 0.602 $\pm$ 0.035 \\
        \bottomrule
    \end{tabular}
\end{table}

\begin{table}[hbt]
    \centering
    \caption{Test set performance comparison across models for Joint Attention detection.}
    \label{tab:results-ja}
    \begin{tabular}{lccc}
        \toprule
        \textbf{Metric} & \textbf{Proposed Model} & \textbf{Conv. Baseline} & \textbf{Best LLM} \\ 
        \midrule
        Accuracy        & \textbf{0.776} $\pm$ 0.027  & 0.626 $\pm$ 0.060 & 0.595 $\pm$ 0.011 \\
        Precision       & \textbf{0.795} $\pm$ 0.033  & 0.645 $\pm$ 0.035 & 0.586 $\pm$ 0.011 \\
        Recall          & \textbf{0.745} $\pm$ 0.100  & 0.562 $\pm$ 0.259 & 0.635 $\pm$ 0.017 \\
        F1-score        & \textbf{0.767} $\pm$ 0.045  & 0.589 $\pm$ 0.178 & 0.610 $\pm$ 0.013 \\
        AUC             & \textbf{0.862} $\pm$ 0.013  & 0.681 $\pm$ 0.013 & 0.595 $\pm$ 0.011 \\
        \bottomrule
    \end{tabular}
\end{table}

%

\begin{figure}[htbp]
\centering
\begin{tikzpicture}
\begin{axis}[
    width=\linewidth,
    height=7cm,
    ybar=2pt,
    bar width=8pt,
    enlarge x limits=0.15,
    ymin=0.25, ymax=1.0,
    ylabel={Score},
    symbolic x coords={Accuracy, Precision, Recall, F1, AUC},
    xtick=data,
    ytick={0.3,0.4,0.5,0.6,0.7,0.8,0.9,1.0},
    ymajorgrids=true,
    grid style={dashed, gray!30},
    legend style={
        at={(0.5,-0.18)}, anchor=north,
        legend columns=3,
        /tikz/every even column/.append style={column sep=0.4cm},
        font=\small
    },
    error bars/y dir=both,
    error bars/y explicit,
    error bars/error bar style={line width=0.8pt, black},
    error bars/error mark options={rotate=90, mark size=3pt, line width=0.8pt},
]

\addplot+[fill=blue!60, draw=blue!80!black] table[
    x=metric, y=mean, y error minus=eminus, y error plus=eplus, col sep=comma
]{
metric,mean,eminus,eplus
Accuracy,0.808,0.023,0.018
Precision,0.752,0.038,0.020
Recall,0.915,0.021,0.026
F1,0.825,0.013,0.015
AUC,0.834,0.007,0.013
};

\addplot+[fill=orange!70, draw=orange!80!black] table[
    x=metric, y=mean, y error minus=eminus, y error plus=eplus, col sep=comma
]{
metric,mean,eminus,eplus
Accuracy,0.529,0.006,0.006
Precision,0.513,0.002,0.004
Recall,0.904,0.057,0.037
F1,0.655,0.018,0.009
AUC,0.429,0.012,0.014
};

\addplot+[fill=green!55!black, draw=green!40!black] table[
    x=metric, y=mean, y error minus=eminus, y error plus=eplus, col sep=comma
]{
metric,mean,eminus,eplus
Accuracy,0.605,0.029,0.035
Precision,0.686,0.061,0.059
Recall,0.369,0.040,0.043
F1,0.480,0.039,0.050
AUC,0.602,0.029,0.035
};

\legend{Proposed Model, Conv. Baseline, Best LLM}
\end{axis}
\end{tikzpicture}
\caption{Mutual Gaze (MG) detection performance across models. Bars show the
mean across independent runs; error bars indicate the observed minimum and
maximum values.}
\label{fig:results-mg-chart}
\end{figure}
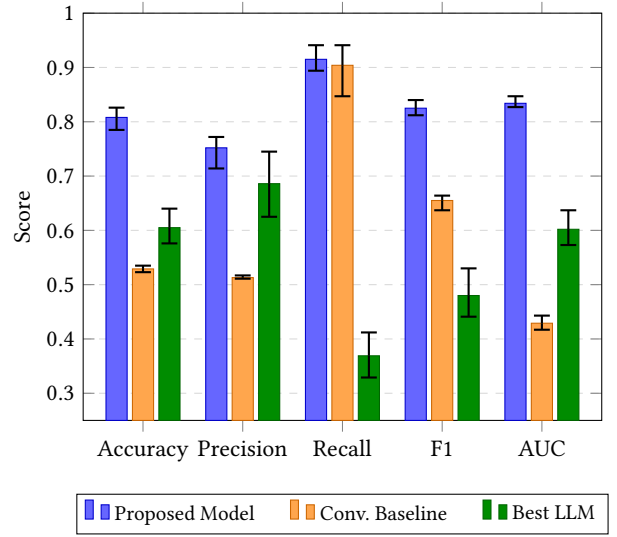

Our proposed model consistently outperforms the baseline architectures, achieving the best balance of precision and recall across both tasks. For Mutual Gaze, the model achieved an F1-score of 0.825 and an ROC-AUC of 0.834. For Joint Attention, it maintained strong performance with an F1-score of 0.767 and an ROC-AUC of 0.862. These results demonstrate a robust discriminative ability to accurately identify target behaviors while maintaining high temporal precision.

In contrast, the lightweight CNN baseline failed to generalize to the held-out test set. For Mutual Gaze, its accuracy reached only 0.529, which is marginally better than random guessing, while its ROC-AUC of 0.429 suggests a failure to learn a reliable decision boundary on the test set. While it performed slightly better on Joint Attention with an AUC of 0.681, it exhibited high variance of $\pm 0.259$ in recall, indicating that a purely convolutional, non-attentional architecture cannot reliably fuse the complex, cross-camera spatial patterns required for interaction analysis.

The Multimodal LLM Baseline showed a strong bias toward over-predicting social events in some contexts while failing to capture them in others. While it achieved a recall for Joint Attention of 0.635, its precision remained low across both tasks (0.686 for MG and 0.586 for JA). Specifically for Mutual Gaze, the LLM’s recall was notably poor at 0.369, resulting in a modest F1-score of 0.480. This behavior suggests that while general-purpose models can identify social cues, they lack the fine-grained sensitivity required to accurately delineate the strict temporal boundaries of MG and JA episodes in a laboratory setting.

\subsection{Computational Efficiency}
Beyond raw classification accuracy, computational efficiency is a critical factor for the real-world adoption of automated coding tools in behavioral research. Psychologists require tools that can process hours of experimental video in a reasonable timeframe. 

When evaluated on an NVIDIA A100 testbed, the proposed model achieved an inference throughput of 9.46 samples per second, despite the pipeline loading all three camera streams while utilizing only two. In stark contrast, the locally served LLM achieved a throughput of just 0.14 samples per second. Furthermore, the LLM required two 40 GB A100 GPUs, whereas our model operated on a significantly smaller footprint. These results demonstrate that our architecture is not only highly accurate at distinguishing complex social cues but also offers the efficiency required for the rapid, scalable post-processing of large-scale psychological datasets.

\section{Discussion}

The results of our evaluation demonstrate that the proposed dual-stream Transformer architecture offers a robust and scalable solution for automating the detection of MG and JA in naturalistic psychological experiments. By directly addressing the limitations of manual behavioral coding, our work bridges a gap between computational modeling and applied behavioral science.

\subsection{Architectural Advantages and Baseline Failures}
The stark performance difference between our model and the baselines highlights the unique challenges of multi-view behavioral modeling. The convolutional baseline's tendency to overfit and its near-random test accuracy suggest that purely local, spatial feature extraction is insufficient for this task. MG and JA are inherently relational constructs; predicting them requires modeling the geometric and semantic relationship between two separate individuals across different camera views. 

Our model succeeds because the pre-trained GazeLLE backbones extract strong, gaze-aware priors, while the Transformer encoder and \texttt{[CLS]} token fusion mechanism effectively map the cross-camera relational dynamics. The results indicate that utilizing a dedicated model, rather than relying on off-the-shelf multimodal LLMs, not only improves classification accuracy but also substantially reduces inference latency and computational overhead.

\subsection{Extensibility and Open Science}
A major challenge in computational behavioral analysis is the domain shift between different laboratory environments. Psychological experiments utilize highly varied physical setups, including different camera angles, focal lengths, lighting conditions, and room geometries. A model trained exclusively on one laboratory's data may experience a drop in accuracy when deployed "out-of-the-box" in another. 

To mitigate this and maximize the utility of our work for the broader scientific community, we are releasing the full model architecture\footnote{\url{https://github.com/child-lab-uj/child-lab-nn}} alongside our pre-trained weights\footnote{\url{https://zenodo.org/uploads/19707957}}, as well as the test dataset\footnote{\url{https://zenodo.org/uploads/19706512}} used to report results in this paper. Because our architecture relies on frozen feature extractors and a relatively lightweight fusion and classification head, it is highly parameter-efficient to train. Other research groups can leverage our pre-trained weights to fine-tune the model on their own specific laboratory configurations using only a small sample of locally annotated data. This dramatically lowers the barrier to entry, allowing labs to adapt the model to their unique ecological constraints and significantly accelerate their research pipelines.

\subsection{Limitations}
Despite its strong performance, our approach has several limitations. First, the pipeline relies on an upstream face detection model (RetinaFace) to generate head bounding boxes. Consequently, any failure in the detection stage cascades down, preventing the model from classifying that frame. 

Second, our current temporal resolution of 1 frame per second (1 fps), chosen to balance computational load and behavioral dynamics, may miss highly ephemeral micro-behaviors that occur on a sub-second timescale. Finally, while the deliberate variation in camera placement during our sessions improved model robustness, the dataset itself is cons  trained to 13 caregiver-infant dyads. Expanding the dataset to include a wider demographic variety would further validate the model's generalizability.

\subsection{Safe and Responsible Innovation Statement}
The automated analysis of human behavior, particularly involving minors in psychological studies, necessitates careful ethical consideration. All data utilized in this study was collected with informed parental consent under approved institutional protocols. Because deep learning models can inadvertently encode demographic biases present in their training data or upstream pre-trained weights, researchers must remain vigilant when deploying these tools across diverse populations. We release our model as an assistive tool to augment, rather than entirely replace, human scientific judgment.

\section{Conclusion}
In this paper, we introduced a novel, highly efficient deep learning architecture for the automated detection of Joint A ttention and Mutual Gaze from dual-camera laboratory setups. By leveraging frozen gaze-aware backbones and a custom token fusion strategy, our model significantly outperforms both CNN and LLM baselines in accuracy, precision, and computational throughput. Supported by a custom annotation tool and an ecologically valid dataset, this work provides behavioral scientists with a scalable mechanism to accelerate interaction research. By open-sourcing our model and pre-trained weights, we provide a parameter-efficient foundation that can be easily fine-tuned by other laboratories, lowering the technical barrier to adopting automated behavioral analysis. Future work will explore end-to-end architectures that bypass the need for explicit face detection and extend the temporal resolution to capture sub-second social dynamics.

\subsection{Future Work}
Building upon this foundation, several promising directions could further enhance the robustness and accuracy of automated Joint Attention detection:

\begin{itemize}
    \item \textbf{Temporal Smoothing:} Incorporating temporal smoothing mechanisms, such as a moving-window voting scheme (e.g., aggregating predictions across adjacent frames), could significantly reduce frame-level noise and improve the temporal stability of the predicted behavioral episodes.
    \item \textbf{Dataset Diversity:} Expanding the dataset to encompass a wider variety of laboratory environments, camera topologies, and demographic groups will be vital for improving the model's out-of-distribution robustness and generalizability.
    \item \textbf{Identity Tracking:} Integrating explicit participant tracking and identity verification would allow the model to reliably distinguish between the parent and child regardless of their position in the room. This would move the system beyond relying on fixed camera assignments, paving the way for the analysis of entirely unconstrained interactions.
    \item \textbf{Multi-Camera Scalability:} Extending the current dual-stream architecture to process an arbitrary number of synchronized camera views. This would allow the system to handle more complex, multi-party interactions and resolve visual occlusions more effectively.
    \item \textbf{Performance Optimization and Tooling:} Pushing the model's classification performance further through architectural refinements, coupled with the development of enhanced, automated tooling for data preparation to streamline the pipeline from raw experimental recordings to model-ready inputs.
\end{itemize}


\begin{acks}
This work was supported by funds from the Polish Ministry of Science and Higher Education assigned to AGH University of Krakow, as well as by the Strategic Program Excellence Initiative at the Jagiellonian University. Computational resources were provided by the PLGrid HPC infrastructure (ACK Cyfronet AGH, grant no. PLG/2025/018713).
\end{acks}

\bibliographystyle{ACM-Reference-Format}
\bibliography{bibliography} 

\end{document}